\def\year{2021}\relax
\documentclass[letterpaper]{article} 
\usepackage{aaai21}  
\usepackage{times}  
\usepackage{helvet} 
\usepackage{courier}  
\usepackage[hyphens]{url}  
\usepackage{graphicx} 
\usepackage{natbib}  
\usepackage{caption} 
\usepackage{amsmath}
\usepackage{booktabs, multirow, makecell}
\usepackage[ruled, vlined]{algorithm2e}
\title{Learning to Retrieve Entity-Aware Knowledge and Generate Responses with Copy Mechanism for Task-Oriented Dialogue Systems}

\author{
  Chao-Hong Tan$^1$\thanks{Equal contribution.}, 
  Xiaoyu Yang$^2$\footnotemark[1], 
  Zi’ou Zheng$^2$\footnotemark[1], 
  Yufei Feng$^2$\footnotemark[1], 
  Tianda Li$^2$\footnotemark[1], 
  Jia-Chen Gu$^1$, \\
  Quan Liu$^{1}$, 
  Dan Liu$^1$, 
  Zhen-Hua Ling$^1$\thanks{Corresponding authors.}, 
  Xiaodan Zhu$^2$\footnotemark[2] \\
  $^1$National Engineering Laboratory for Speech and Language Information Processing, \\
      University of Science and Technology of China, Hefei, China \\
  $^2$ECE \& Ingenuity Labs, Queen's University, Kingston, Canada \\
{\tt \{chtan, gujc, danliu\}@mail.ustc.edu.cn}, {\tt \{zhling, quanliu\}@ustc.edu.cn}\\ 
{\tt \{17xy44, ziou.zheng, feng.yufei, tianda.li, xiaodan.zhu\}@queensu.ca}
}
\begin{document}

\maketitle

\begin{abstract}
Task-oriented conversational modeling with unstructured knowledge access, as track 1 of the 9th Dialogue System Technology Challenges (DSTC 9), requests to build a system to generate response given dialogue history and knowledge access. This challenge can be separated into three subtasks, (1) knowledge-seeking turn detection, (2) knowledge selection, and (3) knowledge-grounded response generation.
We use pre-trained language models, ELECTRA and RoBERTa, as our base encoder for different subtasks. For subtask 1 and 2, the coarse-grained information like domain and entity are used to enhance knowledge usage. For subtask 3, we use a latent variable to encode dialog history and selected knowledge better and generate responses combined with copy mechanism. Meanwhile, some useful post-processing strategies are performed on the model's final output to make further knowledge usage in the generation task. 
As shown in released evaluation results, our proposed system ranks second under objective metrics and ranks fourth under human metrics.
\end{abstract}

\section{Introduction}
The traditional task-oriented dialogue systems, which focuses on providing information and performing actions by the given databases(DB) or APIs, often meet the limitation that the DB/API can not cover enough necessary cases. A good enhance can be achieved with lots of relevant domain knowledge in the form of descriptions, FAQs and customer reviews, which we call unstructured knowledge. Track 1 of the 9th Dialogue System Technology Challenges (DSTC 9)~\cite{gunasekara2020overview}, \textit{Beyond Domain APIs: Task-oriented Conversational Modeling with Unstructured Knowledge Access}~\cite{kim2020domain}, aims at generating a response based on dialogue history and unstructured knowledge access. The whole task can be divided into three subtasks, \textit{knowledge-seeking turn detection}, \textit{knowledge selection} and \textit{knowledge-grounded response}. Test set of this track includes \textit{seen} and \textit{unseen} parts. The unseen test set are collected on different domains, entities, and locales, aiming to evaluate models' generalization ability. 

\textbf{Knowledge-seeking turn detection}, as the first subtask, needs to determine whether the related knowledge is contained in the unstructured knowledge base. In other words, this subtask can be modeled as a binary classification problem. If the model predicts that there exists related knowledge, then subtask 2 (knowledge selection) will search for the most relevant knowledge snippets and then pass them to the generation process (knowledge-grounded response generation). If the model predicts that there is no related knowledge for the specific question, the remaining two subtasks will not be performed.
In this paper, we first conduct an entity matching for each question and then add the domain label from matching results to the end of dialogue history as model input.

\textbf{Knowledge selection} is to retrieve the most relevant knowledge snippets from the database according to the dialogue history and provide information for the subsequent response generation. The dialogue history is a conversation between the human speaker and the machine. Close to the end of the conversation, the human speaker brings up a question about a certain place (e.g., hotel, restaurant) or service (e.g., train, taxi).
The given knowledge database consists of question-answer pairs involving diverse facts and is organized by different domains and entities.
In this paper, we first apply retrieval techniques to narrow down the searching space and then use a neural network initialized by a pre-trained model to formulate the ranking function.

 
\textbf{Knowledge-grounded response generation} requests to give a response automatically from the model using dialogue history and unstructured knowledge as input. There are two different types of dialogue systems, retrieval-based system, and generation-based system. Retrieval-based dialogue system, giving responses from a list of candidate sentences, only has fixed answer forms in candidate sets. To deal with our problem, which needs more flexible and natural responses, the generation-based model is a better choice. Dialogue generation requires an encoder to represent the input and a decoder to generate the response. The network often needs to minimize the cross-entropy loss between the output and the ground truth.
In this paper, we use a latent variable to encode dialog history and selected knowledge better and generate responses combined with copy mechanism.


As shown in released evaluation results~\cite{gunasekara2020overview}, our proposed system ranks second under objective metrics and ranks fourth under human metrics. In the following sections, we will explain the details of our proposed model. Experiment results will be shown next with some analysis and conclusions.

\section{Related Work}
All three subtasks use the pre-trained language model to represent sentences better and deal with the unseen test set. For subtask 1, we use ELECTRA~\cite{Clark2020ELECTRA} as our baseline model, while for subtask 2 and subtask 3, we use RoBERTa~\cite{liu2019roberta} as base encoder. ELECTRA and RoBERTa are BERT-like~\cite{devlin2019bert} architecture with the bi-directional attention mechanism, while GPT~\cite{radford2018gpt, radford2019gpt2} only has uni-directional attention.

\subsubsection{Knowledge-seeking Turn Detection}
To our best knowledge, the Knowledge-seeking Turn Detection is newly proposed by this contest~\cite{kim2020domain}. 
Essentially, Knowledge-seeking Turn Detection is a general classification task that has been explored by NLP community for decades. There are some relevant research topics, such as sentiment analysis and natural language inference (NLI). Currently, the main stream pre-trained models~\cite{devlin2019bert,liu2019roberta,Clark2020ELECTRA} can get state-of-the-art performance on these classification tasks. According to our experiments, ELECTRA reached the highest performance on subtask1, so we select ELECTRA as our baseline model on this subtask.

\subsubsection{Knowledge Selection}
Information retrieval (IR) techniques are widely applied to search for related candidates in retrieval-based knowledge-grounded system. Some researchers ~\cite{song2018ensemble,dinan2018wizard} compute the traditional \textit{tf-idf} score to search the most relevant document to the user's query, while others~\cite{yan2018coupled,zhao2019document,gu-etal-2019-dually,gu-etal-2020-filtering} leverage the power of neural networks to learn the ranking score directly through an end-to-end learning process. Recently, due to the significant improvements on numerous natural language processing tasks, large scale pre-trained language models have also been applied to better model the semantic relevance in knowledge selection~\cite{zhao2020knowledge}. 


\subsubsection{Dialogue Generation}
Two architectures are often used in generation, sequence-to-sequence (Seq2Seq)~\cite{vinyals2015neural, serban2016building, vaswani2017attention} and language model~\cite{radford2018gpt, radford2019gpt2}. Pre-trained language models make a great progress on dialogue generation~\cite{dong2019unified, DBLP:journals/corr/abs-1901-08149, zhang2019dialogpt}. PLATO~\cite{bao2020plato} and PLATO-2~\cite{bao2020plato2} use uni- and bi-directional processing to further pre-train on large-scale Reddit and Twitter conversations dataset to reduce data distribution gaps. Moreover, a latent variable $\mathit{z}$ is used to capture one-to-many relations of post-response pairs.

\subsubsection{Conditional Dialogue Generation}
Being viewed as conditional dialogue response generation, this subtask is closed to persona-chat~\cite{zhang2018personachat}, which aims to generate responses within dialogue by given personality of speakers.~\cite{DBLP:journals/corr/abs-1901-08149} use GPT-2 as their based model by concatenating the personality and dialogue history splitting by speaker tokens. They also use an auxiliary task, binary classification, to decide whether the response is true under given condition. Baseline given by organizer~\cite{kim2020domain} also using GPT-2 with concatenating the knowledge and dialogue history with speaker tokens. 

\subsubsection{Pointer Network}
Knowledge-based dialogue generation struggles with out-of-vocabulary (OOV) words since knowledge will be exact time or some strange proper nouns, which will not be seen by pre-trained language models. To deal with OOV problems,~\cite{see2017point} provided a method to generate words by adding the attention distribution into standard decoder output. It is also called the copy mechanism since the probability of input words would be copied as answers.

\section{Methodology}
Following the task introduction, we separated our system into three subtasks, and three different models were applied respectively.

To clarify the problem, let $S=\{\textbf{s}_1, \textbf{s}_2, ..., \textbf{s}_n\}$ denote the dialogue history, where $\textbf{s}_i$ is the $i^{th}$ utterance in the dialogue and $n$ is the total number of the utterances. 
And let $\textbf{r}=\{r_1, r_2, ..., r_l\}$ denote the response of dialogue, where $r_i$ is the $i^{th}$ token in the dialogue and $l$ is the total number of the tokens. 
The knowledge database is a collection of knowledge snippets that are question-answer pairs to provide certain facts. Each knowledge snippet in the database contains a domain label $dom$, which indicates the domain (e.g., hotel, restaurant, train, taxi) of the knowledge snippet, an entity label $ent$, which indicates the name of the place (e.g., Avalon Hotel) associated with the knowledge snippet and the knowledge document $doc$ which is a question-answer pair  (e.g., Question: Are pets allowed on site? Answer: Not allowed.). 



\subsection{Knowledge-seeking Turn Detection}
For knowledge-seeking turn detection, model needs to determine label $y \in \{0,1\}$ ($1$ indicates the related knowledge exists) based on the dialogue history $S$.  
\subsubsection{Baseline Model}
The baseline in \citet{kim2020domain} is one-step classification. Specifically, the whole given dialogue history $S$ will be fed into GPT-2~\cite{radford2019gpt2} model to get results.
\subsubsection{Knowledge-aware ELECTRA}
The main drawback of GPT-2 approach is that the model fails to consider the knowledge that will be used in subsequent subtasks. This subtask's target is to find if there is a relevant knowledge snippet in the given knowledge base. If knowledge-seeking turn detection fails to detect the knowledge, subsequent subtasks will be missed.

To make use of knowledge, we propose a knowledge-aware ELECTRA model.
The knowledge-aware ELECTRA model leverages knowledge in two steps. Firstly, we conduct an entity matching for each question. 
The details of entity matching will be explained in the retrieval model in Section~\ref{sec:entity_matching}.
After entity matching, if there exists an entity in the knowledge base that also exists in the question, e.g. ``Allenbell hotel’’, we will concatenate the domain label $dom$ to the end of dialogue history $S$.
The input of subtask1 can be format as:
\begin{eqnarray}\label{eq:task1_input}
    \langle bos \rangle \ \textbf{s}_1, \textbf{s}_2, ..., \textbf{s}_{n-1} \ 
    \langle sep  \rangle \ \textbf{s}_n \ dom \ \langle eos  \rangle. 
\end{eqnarray}
Secondly, we add a one-bit knowledge flag to the end of the final hidden vector that corresponds to the token $\langle bos \rangle$ obtained from the ELECTRA model, and they will be fed to the final classifier together.
If the knowledge entity is found, we set the knowledge flag as $1$; otherwise, the knowledge flag was set as $0$. 
In this way, the model will also consider the existence of domain and knowledge. Specifically, there are five domain names used as $dom$ in subtask1 i.e., taxi, train, hotel, restaurant, other. The introduction of domain sequence not only deals with out of domain cases but also makes use of the fact that different domain has different question style.

\subsection{Knowledge Selection}
Knowledge selection aims to select the most plausible knowledge snippets in database according to the dialogue history to facilitate subsequent dialogue completion. In this part, we only deal with the dialogues whose related knowledge snippets are available in the given knowledge database based on knowledge-seeking turn detection judgments.

We propose two base models for the knowledge selection subtask, and the final ensemble model combines the predictions of different base models to improve the selection performance. 
The \textit{Retrieve \& Rank} model first gathers the knowledge snippets of potentially relevant entities from the knowledge base, and then a ranking model is trained to select the most plausible knowledge snippet from the retrieved candidates.
Different from the \textit{Retrieve \& Rank} model, \textit{Three-step} model divides the ranking model into three cascade parts to rank domain, entity and document respectively in order to force the model to take the knowledge hierarchy into account.
We also ensemble these two models together and experiments show the ensemble model has a better performance than each of the two base models. 




\paragraph{Retrieval Model}\label{sec:entity_matching}
The knowledge database for training and validation contains more than 2900 knowledge snippets. Encoding all the knowledge snippets during training is inefficient due to the limitation of memory and computing resources. Given the dialogue history and the ground-truth knowledge snippet, the baseline model in \citet{kim2020domain} re-formulates the task as a binary classification problem during training by randomly sampling one negative knowledge snippet from the database. During evaluation, the model goes through all snippets in the database. 

The baseline model is inefficient during evaluation and takes a much longer time compared with training. The model also suffers from the training-evaluation discrepancy because during training, the model has limited access to the knowledge database comparing to the evaluation. 
To remedy these issues, we proposed a retrieval module at the top of the ranking model to narrow down the scope of potentially useful knowledge snippets in the database, and the retrieved knowledge snippets are used for both training and evaluation.

The retrieval model collects all knowledge snippets whose entity label $ent$ is relevant to the dialogue history $S$. It operates on the entity level, so all knowledge snippets about the same place or service will be / not be collected by the retrieval model at the same time. We show the retrieve algorithm in Algorithm~\ref{alg:algorithm}. 
For each entity label $ent$ in the knowledge base, we check whether it is relevant to the current dialogue. Given an entity name $ent$, we first handle special tokens in the name and generate multiple aliases $\{ent'_1, ent'_2, ..., ent'_m\}$. For example, we generate alias `A and B' for the name `A \& B'.  We then check whether one of the aliases appear in the complete dialogue $S$ using exact match. We say $ent$ matches the dialogue history if at least one alias appears in the history. 

We also perform fuzzy match and check whether alias $ent'$ appears in the last 5 utterances $S'=\{\textbf{s}_{n-4}, ..., \textbf{s}_n\}$. Specifically, we tokenize the $ent'$ with \textit{spaCy}~\cite{spacy} and count the number of tokens with a fuzzy ratio above the threshold $\tau=0.8$ against the utterance. We define the fuzzy match score as the percentage of the tokens that counts. To prevent the case where the fuzzy match method returns too many matched entities, we sort the entity names by the maximum fuzzy match score over all aliases and only include the top 2 entity names.

\begin{algorithm}[t]
 \caption{Retrieval Model}
 \label{alg:algorithm}
\SetAlgoLined
$S~=\{\textbf{s}_1, \textbf{s}_2, ..., \textbf{s}_n\}$ \\
$S'=\{\textbf{s}_{n-4}, ..., \textbf{s}_n\}$ \\
$E~= Alias(ent) =\{ent'_1, ent'_2, ..., ent'_m\}$ \\
\KwResult{$C = Retrieve\_Model(E, S, \tau)$}
$C= [\phantom{g}]$;\\
\For{$ent' \in E$}{
    \For {$\textbf{s} \in S$}{
    \If{$Exact\_Match(ent', \textbf{s})$}{
       $C = C \cup [ent]$;
       }
    }
    $C'= [\phantom{g}]$\\
    \For {$\textbf{s} \in S'$}{
    \If {$ Fuzzy\_Match(ent, \textbf{s}) > \tau$}{
       $C' = C' \cup [ent]$;
       }
    }
   $C' = sort(C')$
}
$C = C \cup C'[:2]$\\
\end{algorithm}

Once an entity name $ent$ matches the dialogue history, we include all knowledge snippets in the database with the same entity name as our candidate knowledge snippets.

\paragraph{Ranking Model}
The ranking model selects the most appropriate knowledge snippet from the candidates produced by the retrieval model. Following \citet{kim2020domain}, we build the ranking model by finetuning the pre-trained RoBERTa-large model~\cite{liu2019roberta}. We concatenate the dialogue history $S$ and the knowledge snippet as the input:
\begin{eqnarray}\label{eq:task2_input}
   X =  \langle bos \rangle \ S \  \langle sep  \rangle \ dom  \ 
    \langle sep  \rangle \ ent \  \langle sep  \rangle \ doc \  \langle eos  \rangle.
\end{eqnarray}
Here $  \langle bos  \rangle,   \langle sep  \rangle,  \langle eos  \rangle$ are special separating tokens pre-defined in RoBERTa. Based on the final layer hidden vector that corresponds to the token $ \langle bos  \rangle$, the ranking model outputs a ranking score $p$.
\begin{eqnarray}
   l_{ \langle bos  \rangle} = Roberta(X)\\
   p = f(l_{ \langle bos  \rangle})
\end{eqnarray}
where $f$ is a one-layer classifier. We pick the one with the highest ranking score as our final prediction for knowledge selection. 


\paragraph{Three-step Model}
\textit{Three-step} model is similar to the ranking model, but it is applied to domain level, entity level and document level respectively in a pipeline rather than to the document level directly. The input format of the document submodule is the same as Equation~(\ref{eq:task2_input}), and the input format of domain and entity submodules are:
\begin{eqnarray}
   \text{domain:} & \langle bos \rangle \ S \  \langle sep  \rangle \ dom \ \langle eos  \rangle.   \\
   \text{entity:} & \langle bos \rangle \ S \ \langle sep  \rangle  \ dom \ \langle sep  \rangle \ ent \ \langle eos  \rangle. 
\end{eqnarray}
The candidates of entity submodule are only from the most plausible domain and the candidates of document submodule are only from the most plausible entity. \textit{Three-step} model is built beyond RoBERTa-base model~\cite{liu2019roberta}.

\paragraph{Finetuned with Data Augmentation}
We generate more dialogues using the question-answer pair in the knowledge database to enhance the training data, and we finetune the domain and entity submodules in the \textit{Three-step} model with the combination of original training data and generated dialogues. Specifically, for each entity, we first sample the dialogue length according to the number of question-answer pairs under this entity and then we randomly sample the question-answer pairs and concatenate them together as a new dialogue. To mimic the situation of topic shift in the real scenario, we generate another part of dialogue under another entity and concatenate them together eighty percent of the time.

\paragraph{Ensemble Model}
We ensemble the \textit{Retrieve \& Rank} model with \textit{Three-step} model together. The ensemble model is also divided into three submodules. In each submodule, we take the prediction with higher probability as results. 

\subsection{Knowledge-grounded Response Generation}

\begin{figure}[t]
\centering
\includegraphics[width=.95\columnwidth]{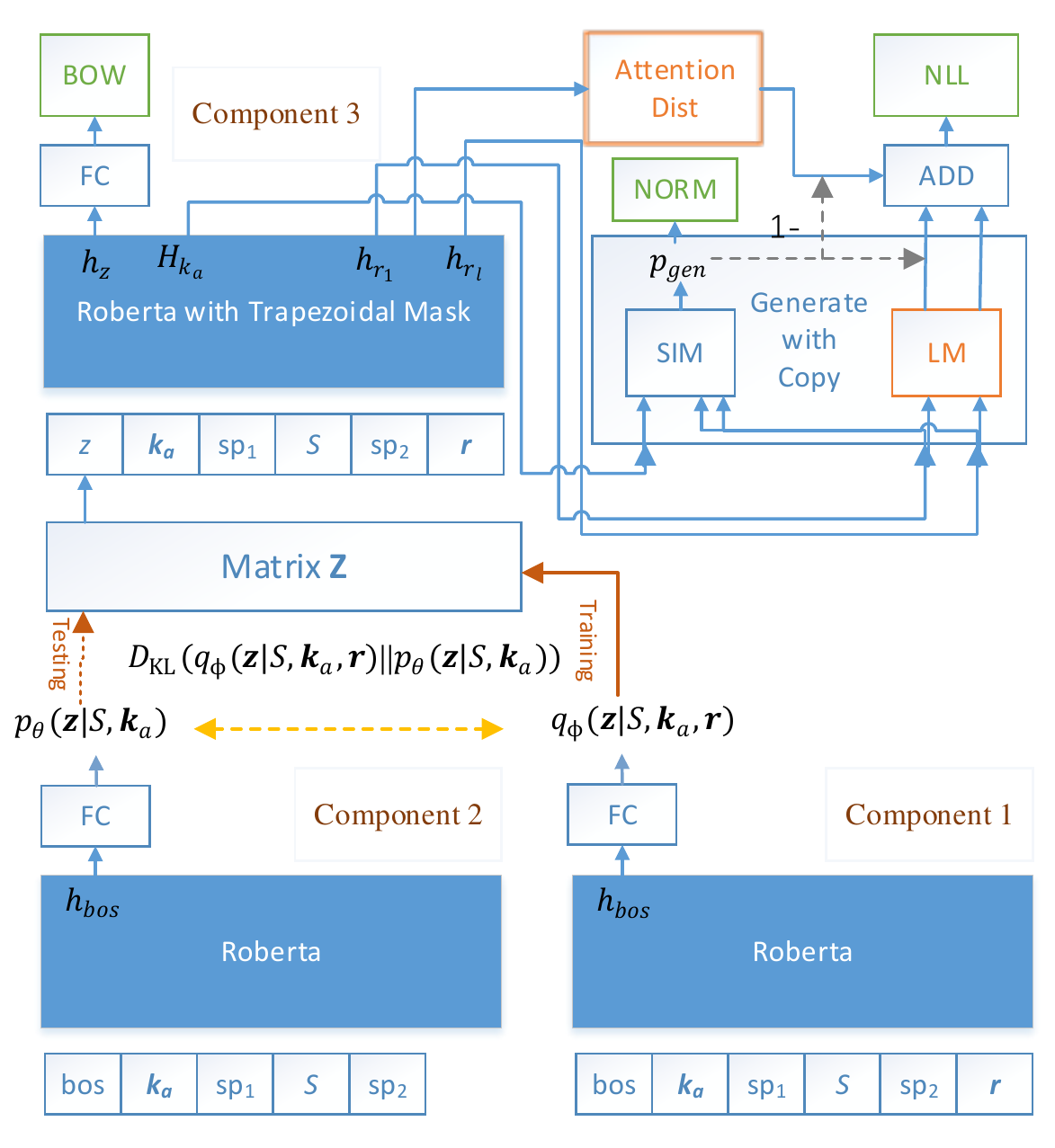} 
\caption{The architecture of subtask-3 model.}
\label{figure3_1}
\end{figure}

\subsubsection{Basic Architecture} 
Inspired by PLATO model~\cite{bao2020plato,bao2020plato2}, we notice that the latent variable may be useful for getting the relations among dialogue history $S$, answer part of knowledge $\textbf{k}_a$, and response $\textbf{r}$. We first concatenate them together with some special tokens:
\begin{equation}
    \langle bos \rangle \ \textbf{k}_a \  \langle sp_1  \rangle \ \textbf{s}_1 \
    \langle sp_2  \rangle \ \textbf{s}_2 \text{...} \langle sp_2 \rangle \ \textbf{r} \  \langle eos   \rangle. 
\end{equation}
where $\langle sp_1  \rangle$ represents the first speaker, $\langle sp_2  \rangle$ represents the second speaker in our two speakers' dialogue.
Following the setting in PLATO, we assume that the latent variable $\mathit{z}$ is one K-way categorical variable. We want to encode response information into the $\mathbf{Z}$ matrix, and each row of the matrix represents a special $\mathit{z}$ corresponding to given examples. We estimate posterior probability $q_{\phi}(\mathbf{z}|S, \textbf{k}_a, \textbf{r})$ which are used to select $\mathit{z}$ from matrix $\mathbf{Z}$.

However, $\mathit{z}$ can not be calculated using the posterior probability at test time without accessing responses. Thus we use prior probability $p_{\theta}(\mathbf{z}|S, \textbf{k}_a)$ to approach the posterior probability.
Inspired by VAE~\cite{kingma2014VAE}, we reduce the KL-divergence (KLD) loss
\begin{equation}
    l_{KLD} = D_{KL}(q_{\phi}(\mathbf{z}|S, \textbf{k}_a,\textbf{r})||p_{\theta}(\mathbf{z}|S, \textbf{k}_a))
\end{equation}
to minimize the gap between posterior and prior probability.
Thus we split our model into three components, as shown in \ref{figure3_1}. We use the trapezoidal mask to prevent response information leakage from our bi-directional encoder.
The first component calculates the posterior probability $q_{\phi}(\mathbf{z}|S, \textbf{k}_a,\textbf{r})$ with responses as input. At the same time, the matrix $\mathbf{Z}$ also will be updated to get the information of response. The second component is to calculate the prior probability $p_{\theta}(\mathbf{z}|S, \textbf{k}_a)$ without response. After obtaining the posterior and prior probability of $\mathit{z}$, we calculate KLD loss.

With dialogue history, knowledge and $\mathit{z}$, the third component generates responses with reducing negative log-likelihood (NLL) loss,
\begin{equation}
\begin{split}
l_{NLL} &= -{E}_{\mathit{z}\sim q_{\phi}(\mathbf{z}|S, \textbf{k}_a,\textbf{r})} \log{p(\textbf{r}|S, \textbf{k}_a,\mathit{z})} \\
    &= -{E}_{\mathit{z}\sim q_{\phi}(\mathbf{z}|S, \textbf{k}_a,\textbf{r})} \sum_{t=1}^T \log{p(r_t|S, \textbf{k}_a,\mathit{z}, \textbf{r}_{<t})}
\end{split}
\end{equation}
The bag-of-words (BOW) loss~\cite{zhao-etal-2017-learning} is also employed to facilitate the training process of latent discrete variables,
\begin{equation}
\begin{split}
    l_{BOW} &= -{E}_{\mathit{z}\sim q_{\phi}(\mathbf{z}|S, \textbf{k}_a,\textbf{r})} \sum_{t=1}^T \log{p(r_t|S, \textbf{k}_a,\mathit{z})} \\
            &= -{E}_{\mathit{z}\sim q_{\phi}(\mathbf{z}|S, \textbf{k}_a,\textbf{r})} \sum_{t=1}^T \log{\frac{e^{f_{r_t}}}{\sum_{v \in V}{e^{f_v}}}}
\end{split}
\end{equation}
where $V$ refers to the whole vocabulary, and $f$ is a linear layer with softmax on the hidden output of $\mathit{z}$, $\boldsymbol{h_z}$, which tries to predict the words of response without position information,
\begin{equation}
    f = \mathrm{softmax}(\mathbf{W_1}\boldsymbol{h_z} + \boldsymbol{b_1})
\end{equation}
We only pass through component 2 and 3 at test time.

\subsubsection{Knowledge Copy}
Without the ability to utilize knowledge directly, the pure dialogue generation model will only focus on grammatical/semantical correctness, naturalness, and appropriateness. However, another main target of this task, relevance to the given knowledge, has not been taken into account. Motivated by pointer-generator network~\cite{see2017point}, we propose the \textit{Knowledge Copy} method to calculate the probability of response $p(\textbf{r}|S, \textbf{k}_a,\mathit{z})$, denoted as $\mathbf{P}(V)$ for convenience, by combining generation distribution $\mathbf{P_{lang}}(V)$ with the knowledge-attention distribution $\mathbf{P_{att}}(V')$. $V$ is the whole vocabulary, and $V'$ is the words appearing in the knowledge.

We calculate $\mathbf{P_{lang}}(V)$ at each time step $t$, noted as $\mathbf{P_{lang}}(V, t)$. It is the output of the decoder, which consists of a dense layer with activation function GELU~\cite{hendrycks2016gaussian}, and a linear projection layer to vocabulary size,
\begin{equation}
\begin{split}
    \boldsymbol{d_t} &= \mathrm{GELU}({\mathbf{W_2}\boldsymbol{h_t} + \boldsymbol{b_2}})  \\
    \mathbf{P_{lang}}(V, t) &= \mathrm{softmax}({
        \mathbf{W_3}\boldsymbol{d_t} + \boldsymbol{b_3})
    }.
\end{split}
\end{equation}
where $h_t$ means hidden output of the encoder at current time step.

To obtain $\mathbf{P_{att}}(V')$, we extract the last attention layer of RoBERTa and calculate the average value through multi-heads to get the representation of knowledge-attention distribution. The attention distribution includes knowledge, context and response parts, but what we exactly need is the part from response to knowledge. So we extract that part and normalize to make sure the distribution under probability's limitation. 

The gate at each time step $p_{gen}(t) \in (0, 1)$, which controls how many information can be generated, is decided by the knowledge and hidden outputs of the encoder,
\begin{equation}
    p_{gen}(t) = \sigma(\mathbf{W_p}[\boldsymbol{k_m} \circ \mathbf{h_t}, \mathbf{k_m}, \mathbf{h_t}] + b_p)
\end{equation}
And knowledge representation $\boldsymbol{k_m}$ can be calculated by
\begin{equation}
    \boldsymbol{k_m} = \frac{1}{N_k}{\sum_{i=1}^{N_k}{\boldsymbol{h_{k_i}}}}
\end{equation}
where $N_k$ is the token numbers of knowledge in one example, and $\boldsymbol{h_{k_i}}$ is the top hidden outputs in knowledge part.

The total probability used to generate our response is the sum of two distributions with $p_{gen}(t)$ as weight,
\begin{equation}
    \mathbf{P}(V, t) = p_{gen}(t)\mathbf{P_{lang}}(V, t) + (1 - p_{gen}(t))\mathbf{P_{att}}(V', t) \label{eq:t3-1}
\end{equation}
where $\mathbf{P}(V, t)$ can be viewed as $p(r_t|S, \textbf{k}_a,\mathit{z}, \textbf{r}_{<t})$ to calculate $l_{NLL}$. If $w \notin V$ is an out-of-vocabulary (OOV) word with $P_{lang}(w, t)$ equaling or approaching to zero, it still can be generated if it appears at knowledge part, which means $P_{att}(w, t)$ is not zero. This architecture is shown as \textit{generate with copy} part in figure \ref{figure3_1}.

The knowledge-copy mechanism provides an efficient way to generate sentences under the given knowledge, reducing the pressure added on the decoder. It makes the model learn how to \textbf{directly} use knowledge, which is easier to generalize to unseen knowledge. More analysis will be done in later sections. Due to the  benefits of knowledge-copy mechanism, we want to use it as much as possible, that is, to make $p_{gen}(t)$ smaller. So a normalization loss will be added to the final loss, which is
\begin{equation}
    l_{norm} = \sum_{t=1}^T{{p_{gen}}^2(t)}
\end{equation}
where $T$ means the maximum time step of response generation.

In summary, the total integrated loss of our generation model is
\begin{equation}
    l_{total} = \lambda_1 l_{NLL} + \lambda_2 l_{BOW} + \lambda_3 l_{KLD} + \lambda_4 l_{norm}
\end{equation}

\subsubsection{Segmented Response Generation}
In this subtask, one response can be viewed as two parts. One is knowledge-response which needs lots of information from the knowledge candidate, and the other is greeting-response, which is less-information and does not request so much knowledge background. With given dialogue history, knowledge and latent variable $\mathit{z}$, these two parts can be generated by different model respectively due to a reasonable assumption of conditional independence. We use this segmented response generation (SRG) method in our experiments and show its power in the analysis part. Note that the knowledge-copy mechanism is not applied for greeting-response generation.

\subsubsection{Modified Beam Search}
Since beam search may yield similar sentences and diversity beam search~\cite{vijayakumar2018diverse} may generate lots of useless sentences, we propose the first-word-fixed beam search (FFBS) for our generation. Concretely, the first words of the generated responses are selected by top-k probability and fixed as different groups, and regular beam search will be done separately in each group. We fix the first words because they almost control the whole sentence development in auto-regressive generation models.

\subsubsection{Post-processing Strategies}
Response with more knowledge is better in most cases. Inspired by this, post-processing (PP) is performed in generation using similarity between response and knowledge to re-rank our candidate responses generated by FFBS. BERTScore~\cite{zhang2020bertscore} is applied to calculate the semantic similarity between knowledge-response and the knowledge candidate. However, it may score extremely high when the two sentences are the same. To ensure the flexibility of responses, we also calculate the Jaro–Winkler distance (JWD)~\cite{Jaro1989Advances, Winkler1990String}, a kind of edit distance
to evaluate the sentence similarity.
To sum up, the score we use to re-rank response the candidate is 
\begin{equation}
    S_{total} = \mu_1 S_{NLL} + \mu_2 S_{BERT} - \mu_3 S_{JWD}
\end{equation}
where $S_{NLL}$ equals to log probability of each beam output from the generator, which will also be normalized in $[0, 1]$. The sentences with the highest $S_{total}$ through candidates will be selected as our final responses.


\section{Experiment}  
  
\begin{table}[t]
\centering
\begin{tabular}{@{}p{2cm}p{2cm}<{\raggedleft}p{2cm}<{\raggedleft}}
\toprule
    Split        & Total turns& K-turns \\ \midrule
    Train        &  71,348    &  19,184       \\
    Valid        &  9,663     &  2,673        \\
    Test(total)  &  4,181     &  1,981        \\
    Test(unseen) &  --        &  264          \\
    Test(seen)   &  --        &  1,717        \\ \bottomrule
\end{tabular}
\caption{Statistics of dialogues. K-turns means this turns required knowledge.}
\label{table:3-1}
\end{table}

\begin{table}[t]
\centering
\begin{tabular}{@{}p{2cm}p{2cm}<{\centering}p{2.5cm}<{\centering}@{}}
\toprule
    Domain       &  Entities  &  Snippets        \\ \midrule
    Hotel        &  \ 33 / 178  &  1,219 / 4,346   \\
    Restaurant   &  110 / 391 &  1,650 / 7,155   \\
    Train        &  1 / 1     &  5 / 5           \\
    Taxi         &  1 / 1     &  26 / 26         \\
    Attraction   &  \ -- / 97   &  \quad-- / 507        \\ \hline
    Total        &  145 / 668 &  \ 2,900 / 12,039  \\ \bottomrule
\end{tabular}
\caption{Statistics of knowledge, \textit{Train}/\textit{Test}. \textit{Valid} is the same as \textit{training} dataset.}
\label{table:3-2}
\end{table}

\begin{table*}[t]
\centering
\resizebox{.99\linewidth}{!}{
\begin{tabular}{@{}l|c|c|ccc@{}}
\toprule
\multirow{2}{*}{Model} & Knowledge Detection & Knowledge Selection & \multicolumn{3}{c}{Knowledge Grounded Generation}          \\ 
\multicolumn{1}{c|}{}                       & Precision/Recall/F1                      & MRR@5/Recall@1/Recall@5                  & BLEU-1/2/3/4                & METEOR & ROUGE-1/2/L          \\ 
\midrule
Baseline                                    & 0.9933 / 0.9021 / 0.9455                     & 0.7263 / 0.6201 / 0.8772                     & 0.3031 / 0.1732 / 0.1005 / 0.0655 & 0.2983 & 0.3386 / 0.1364 / 0.3039 \\ 
Ours                                        & 0.9933 / 0.9677 / 0.9803                     & 0.9195 / 0.8975 / 0.9460                     & 0.3779 / 0.2532 / 0.1731 / 0.1175 & 0.3931 & 0.4204 / 0.2113 / 0.3765 \\
\bottomrule
\end{tabular}
}
\caption{Objective metrics on test set.}
\label{table:3-3}
\end{table*}

\begin{table*}[t]
\centering
\begin{tabular}{@{}l|ccc|cccc}
\toprule
\multirow{2}{*}{Subtask-2 Model} & \multicolumn{3}{c|}{Validation}      & \multicolumn{4}{c}{Error Number@1}    \\ 
\multicolumn{1}{c|}{}                                  & MRR@5  & Recall@1 & Recall@5 & Domain  & Entity & Document & Total \\ 
\midrule
Roberta (baseline)                                  
&  0.8691      &  0.8058        &    0.9428      
& 215 & 72 & 131 & 418\\

Retrieve \& Rank                 
& \textbf{0.9747} & 0.9622 & \textbf{0.9880}
&16 & 41 & 44 & 101\\
Three-step w/ data aug.                                 
&   0.9739     &   0.9660       &  0.9828        
&  \textbf{7}      &     \textbf{37}     &  47     &  91 \\ 
Three-step w/o data aug.
&   0.9692    &    0.9607      &   0.9783      
&   11     &    43      &    51   &  105 \\ 
Our Ensemble Model                                     
&  0.9743      &   \textbf{0.9678}       &     0.9813     
& 9 & 38 &  \textbf{39} & \textbf{86} \\
\bottomrule
\end{tabular}
\caption{The performance of models on validation set (subtask 2).}
\label{table:subtask-2}
\end{table*}

\begin{table}[t]
\centering
\begin{tabular}{@{}lccc@{}}
\toprule
Model        & Accuracy & Appropriateness & Average \\ \midrule
Baseline     & 3.7155   & 3.9386          & 3.8271  \\ 
Ours         & 4.3793   & 4.2755          & 4.3274  \\ 
Ground-Truth & 4.5930   & 4.4513          & 4.5221  \\ \bottomrule
\end{tabular}
\caption{Human metrics on test set.}
\label{table:3-4}
\end{table}

\begin{table}[t]
\centering
\begin{tabular}{@{}lccc@{}}
\toprule
Subtask-1 Model        & Precision & Recall & F1-score \\ \midrule

Baseline(GPT2)     & 0.957   & 0.966         & 0.961  \\ 
Xlnet              & 0.944   & 0.983         & 0.963 \\
Roberta            & 0.947   & 0.993         & 0.970 \\
ELECTRA            & 0.950   & 0.994         & 0.972 \\ 
Our                & \textbf{0.996}   &  \textbf{0.999}         &   \textbf{0.998}  \\ \bottomrule
\end{tabular}
\caption{Models performance on validation set (subtask1).}
\label{table:Dstc1}
\end{table}

\begin{table}[t]
\centering
\begin{tabular}{@{}lccc@{}}
\toprule
Subtask-3 Model              & BLEU-4   & METEOR & ROUGE-L \\ \midrule
Ours (w/ PP)       & \textbf{0.1450}   & \textbf{0.4526} & \textbf{0.4256}  \\
\quad--SRG        & 0.1366   & 0.4400 & 0.4155  \\ \midrule
Ours (w/o PP)      & \textbf{0.1512}   & 0.4534 & 0.4317  \\ 
\quad--FFBS        & 0.1490   & \textbf{0.4536} & \textbf{0.4342}  \\ 
\qquad--SRG       & 0.1473   & 0.4485 & 0.4311  \\ 
\quad\qquad--Copy  & 0.1424   & 0.4322 & 0.4194  \\ 
\qquad\qquad--Z    & 0.1327   & 0.4210 & 0.4096  \\ \bottomrule
\end{tabular}
\caption{Ablation results under objective metrics on validation set (subtask 3)}
\label{table:3-5}
\end{table}

\begin{table*}[t]
\centering
\begin{tabular}{@{}r|p{.8\linewidth}@{}}
\toprule
Context            & Can I get a written confirmation for the taxi booking? \\ \hline
Knowledge(Body)    & Booking confirmations will be sent via \textbf{text} messages \textbf{shortly}. \\ \hline
Ground-Truth       & You will be sent a confirmation \textbf{text} very \textbf{shortly}. Is there anything else I can help you with? \\ \hline
Ours               & Confirmations will be sent via \textbf{text} messages \textbf{shortly}. Can I help you with anything else? \\ \hline
Ours (w/o PP)      & Yes, you will get a confirmation via \textbf{text}. Can I help you with anything else? \\ \hline
Ours (w/o Copy)    & Yes, you will get a confirmation via \textbf{email}. Is there anything else I can help you with? \\ \midrule[.8pt]
Context            & By the way, what hours is the \textbf{station} where the \textbf{train} departs open? \\ \hline
Knowledge(Body)    & The hours of operation are \textbf{Monday-Saturday 05:00 - 23:00}, and on \textbf{Sundays 07:00 - 22:55}. \\ \hline
Ground-Truth       & \textbf{Monday through Saturday 5 AM to 11:00 PM}, \textbf{Sundays 7 AM to 11:55 PM}. Can I help you with any more information? \\ \hline
Ours               & Hours of operation at the \textbf{train station} are \textbf{Monday-Saturday 05:00-23:00} and on \textbf{Sundays 07:00 - 22:55}. Anything else I can do for you? \\ \hline
Ours (w/o PP)      & The station opens at \textbf{5:00 am Monday-Saturday} and \textbf{7:00 on Sundays}. Anything else I can do for you? \\ \hline
Ours (w/o Copy)    & The station opens at \textbf{5 am Monday-Saturday} and \textbf{7 am on Sunday}. Anything else I can do for you? \\ \bottomrule
\end{tabular}
\caption{Generation examples on validation set (subtask 3).}
\label{table:3-6}
\end{table*}

\subsection{Dataset}
We tested our model on \textit{validation} and \textit{test} dataset provided by this challenge track. The validation dataset, as well as \textit{training} dataset, consist of only seen domain data, which is an augmented version of MultiWoz 2.1~\cite{budzianowski2018multiwoz, eric2019multiwoz} introducing knowledge-seeking turn for dialogue. On the other hand, the \textit{test} dataset consists of two parts, seen domain and unseen domain. The seen domain is an unlabeled test set of the augmented MultiWoz 2.1, while the unseen domain is a new set of unseen conversations collected from scratch also including turns that require knowledge access~\cite{kim2020domain}. The organizer also provides a new knowledge candidate collection for testing. The statistics of dialogues and knowledge are shown in Table \ref{table:3-1}, \ref{table:3-2}.

\subsection{Experiment Details}
Firstly, we initialized our models with pre-trained models' weights obtained from \textit{HuggingFace}'s model hub~\cite{Wolf2019HuggingFacesTS}. And then, our models were fine-tuned on the dataset provided for our special task, respectively.

For subtask 1, we used ELECTRA-base to initialize our model. Learning rate was set to $6.25e\text{-}5$, and batch size was set to $16$.

For subtask 2, RoBERTa-large was used to initialize the \textit{Retrieve \& Rank} model. Learning rate was $1e\text{-}5$ and the batch size was $72$. We first retrieve entities to narrow down the candidate searching scope and then take their corresponding snippets for training and evaluation. Specifically, during training, we randomly sample five negative snippets from the retrieval results. During evaluation, we rank all retrieved snippets to get final predictions.
For \textit{Three-step} model and the ensemble model, we initialized them with RoBERTa-base. Learning rate was $6.25e\text{-}5$, batch size was $4$ and the candidate number in each sample was $6$. For data augmentation, we generated 100 dialogues for each entity in the training/validation knowledge database.

For subtask 3, we used RoBERTa-base to initialize our model, and these hyperparameters were kept the same as the baseline provided by the organizer. Learning rate was set to $6.25e\text{-}5$, and batch size was set to $4$, with $32$ gradient accumulation steps. The number of hidden variable $z$ was set to $5$. To make the knowledge-copy mechanism not merely copy the whole sentence, we masked punctuates of knowledge while getting knowledge-attention distribution. The hyperparameter $\lambda_{1-4}$ and $\mu_{1-3}$ were set to $1$ for convenience.
We trained our model in 10 epochs. 
At the generation stage, we used FFBS to get responses, with $4$ groups and $2$ beams and thus we had $8$ ($4\times2$) responses for one input in total.

All code is published to help replicate our results.\footnote{https://github.com/lxchtan/DSTC9-Track1}

\subsection{Metrics}
Some specific objective metrics are evaluated in each task, including precision, recall, f1-score in subtask 1, mrr@5, recall@1, recall@5 in subtask 2, BLEU-$1$/$2$/$3$/$4$, meteor, rouge-$1$/$2$/L in subtask 3. Human evaluation is also performed in subtask 3, including two aspects, appropriateness and accuracy. Appropriateness metric means how well the response is naturally connected to the conversation, ranging from 1 to 5, and the larger, the better. Accuracy means, with a given reference knowledge, how accurate each system's response is on a scale of 1 to 5, and also the larger, the better.

\subsection{Evaluation Results}
Table \ref{table:3-3} and \ref{table:3-4} present the evaluation results on test dataset of our all three subtasks. Compared with the baseline in \citet{kim2020domain}, our model achieves huge improvement in all three subtasks. In addition, human metrics show that the performance of our model is close to human compared with the ground-truth. The results provided by the organizer~\cite{gunasekara2020overview} show that we rank second under objective metrics and rank fourth under human metrics.

\section{Analysis}

Ablation experiments were conducted to demonstrate the importance of each component in our proposed model, and the results were reported on \textit{validation} set on each subtask.
\subsubsection{Subtask 1}
As shown in Table~\ref{table:Dstc1}, in addition to our model, we report the results of Xlnet, RoBERTa, ELECTRA and GPT-2 on the validation set of subtask 1 as well. The performance of our knowledge-aware ELECTRA outperforms all of these models with a substantial margin. Compared with the original ELECTRA, the F1-score of our model has 2.6\% improvement, which indicates knowledge does help ELECTRA perform better.

\subsubsection{Subtask 2}
We changed the language model from GPT2 to Roberta-base in the model proposed in~\cite{kim2020domain} and used it as our baseline.
The evaluation results of our ensemble model on the test dataset in subtask 2 are shown in Table~\ref{table:3-3}, and it outperforms the baseline by a large margin. 

To have a better idea of the performance of the base models, we present the results of our ensemble model and the two base models (i.e., \textit{Retrieve \& Rank } model and \textit{Three-step } model) in Table~\ref{table:subtask-2}. 
We also show statistics about the number of errors that occur at the domain, entity, and document level. 
Results on the validation set are based on the ground-truth labels of subtask 1.
We can find that \textit{Retrieve \& Rank } model has a better recall@5 while \textit{ Three-step } model with data augmentation provides better recall@1, especially better in domain and entity prediction. The ablation test on data augmentation also shows that the data augmentation technique is helpful at the domain and entity level.
Since the test set involves out-of-domain data, there is a gap between performance on the validation and test set. Further study is required to improve the transferability and generalization of our models.

\subsubsection{Subtask 3}
We used ground-truth labels of subtask 1 and 2 in order to study the components of subtask 3 independently. The result is shown in table \ref{table:3-5}. At the part without post-processing, we can easily find that the latent $\mathit{z}$ contributes about $1\%$ to on each metrics, while knowledge-copy mechanism wins about $0.5\%$ on BLEU-4, about $1.6\%$ on METEOR, about $1.2\%$ in ROUGE-L. SRG and FFBS also improve our model under objective metrics.
Note that the use of post-processing could increase human evaluation scores, while the objective scores may reduce. However, after using SRG, objective scores could rise  without post-processing, as shown in the row \textit{Ours (w/ PP)} and \textit{Ours (w/ PP) --SRG}.
Since objective metrics are not good enough to reflect the accuracy of used knowledge, we select some examples to explore in Table \ref{table:3-6}. We can find that post-processing and knowledge-copy mechanism have a stronger ability to capture information of knowledge.

\section{Conclusion}
This paper describes our overall system that is evaluated in Track 1 of DSTC 9. 
Pre-trained language models, ELECTRA and RoBERTa, are used as our base encoder, and task-specific components are applied to improve performance. In the released evaluation results, we rank second under objective metrics and rank fourth under human metrics. Considering the gap between validation and test set, it is worthwhile for us to further study how to generalize our model in a better way, that is, transferring our in-domain system to the out-of-domain scenario.

\bibliography{ref.bib}
\end{document}